\documentclass[a4paper,conference]{IEEEtran}
\IEEEoverridecommandlockouts
\usepackage{cite}
\usepackage{amsmath,amssymb,amsfonts}
\usepackage{algorithmic}
\usepackage{graphicx}
\usepackage{textcomp}
\usepackage{xcolor}
\usepackage{comment}
\usepackage{algorithm}
\usepackage{booktabs}
\usepackage{multirow}
\usepackage{multicol}
\usepackage{stfloats}
\graphicspath{ {./images/} }
\def\BibTeX{{\rm B\kern-.05em{\sc i\kern-.025em b}\kern-.08em
    T\kern-.1667em\lower.7ex\hbox{E}\kern-.125emX}}
\begin{document}

\title{Improved Deep Classwise Hashing With Centers Similarity Learning for Image Retrieval
}

\author{\IEEEauthorblockN{Ming Zhang}
\IEEEauthorblockA{\textit{Department of Electrical Engineering} \\
\textit{City University of Hong Kong}\\
Hong Kong, China \\
mzhang367-c@my.cityu.edu.hk}
\and

\IEEEauthorblockN{Hong Yan}
\IEEEauthorblockA{\textit{Department of Electrical Engineering} \\
\textit{City University of Hong Kong}\\
Hong Kong, China \\
h.yan@cityu.edu.hk}
}

\maketitle

\begin{abstract}
Deep supervised hashing for image retrieval has attracted researchers' attention due to its high efficiency and superior retrieval performance. Most existing deep supervised hashing works, which are based on pairwise/triplet labels, suffer from the expensive computational cost and insufficient utilization of the semantics information. Recently, deep classwise hashing introduced a classwise loss supervised by class labels information alternatively; however, we find it still has its drawback. In this paper, we propose an improved deep classwise hashing, which enables hashing learning and class centers learning simultaneously. Specifically, we design a two-step strategy on center similarity learning. It interacts with the classwise loss to attract the class center to concentrate on the intra-class samples while pushing other class centers as far as possible. The centers similarity learning contributes to generating more compact and discriminative hashing codes. We conduct experiments on three benchmark datasets. It shows that the proposed method effectively surpasses the original method and outperforms state-of-the-art baselines under various commonly-used evaluation metrics for image retrieval.
\end{abstract}


\section{Introduction}
In the big data era, there is an explosive growth of data uploaded to the Internet every day. How to search the high dimensional data in a large-scale database both efficiently and effectively becomes more important for various social platforms and search engines. Given a query image, the task of content-based image retrieval (CBIR) is finding all the related items to the query in the database according to their visual features.
CBIR has witnessed great progress in the last decades. Suppose both images in the database and the query image are encoded in real-valued features. Thus, to find the closest items in the database to the query, it is realized by ranking the database items by their distance to the query in the feature space. However, for a database with millions of images, the computation and storage cost by using real-valued features are expensive. Moreover, it brings a high burden to the mobile or embedded devices where the resources are limited for both storage and latency.

As a powerful approximate nearest neighbor search technique, hashing has been widely employed for information retrieval. The goal of hashing is mapping high dimensional data into lower dimensional binary codes while preserving their original similarity relations of the input space. Preformed with the binary codes, it takes the advantages of extremely fast retrieval speed and low storage cost. Generally, existing data-dependent hashing methods can be categorized into unsupervised hashing~\cite{weiss2009spectral,gong2012iterative} and supervised hashing~\cite{liu2012supervised,shen2015supervised,gui2017fast}.Unsupervised hashing methods learn hashing functions in different distance measurements without considering the label information. Whereas, supervised hashing incorporates labels information e.g. pairwise labels into the hashing learning process, which is known to outperform the unsupervised hashing with the same code length.

With the great success of deep learning in the past years, deep supervised hashing~\cite{li2016feature,li2017deep,chen2019deep,cao2017hashnet,8759067} proposes to apply deep neural networks to extract feature representations for hashing functions learning. Among them, some works with end-to-end training manners have greatly improved the retrieval performance and shown superior results to traditional supervised hashing methods. Most existing deep supervised hashing utilizes the pairwise label information to examine the similarity between pairs of samples. A representative work is Deep Pairwise Supervised Hashing (DPSH)~\cite{li2016feature}, which minimizes the Hamming distance between each pair of similar samples while maximizing the Hamming distance between each pair of dissimilar samples. The similar samples refer to the samples sharing at least one class and their similarity is denoted as `1'. While dissimilar samples refer to the samples sharing no common object class and their similarity is denoted as `0' or `-1'. Inspired by DPSH, some variations are proposed, e.g., Deep Supervised Hashing with Triplet labels (DTSH)~\cite{wang2016deep} and Deep Supervised Discrete Hashing (DSDH)~\cite{li2017deep}.

The main problem of pairwise/triplet labels-based deep supervised hashing is the tremendous computation cost. The time complexity of the algorithm is $O(n^2)$, where $n$ is the number of samples in the training set. In practice, it is solved by a mini-batch sampling of inputs during training. However, this results in a side effect: the labeled data cannot be fully utilized because some supervised information is always discarded during iterative learning. Consequently, the learned binary hashing codes are suboptimal to preserve the semantics relation of the dataset. Some works have been proposed to overcome this issue, e.g., DAGH~\cite{chen2019deep} utilizes the anchor graph to efficiently connect all the training samples and the anchors under the deep hashing framework. 

Recently, some novel deep hashing methods propose directly use labels information as supervision. One typical example is Deep Classwise Hashing (DCWH)~\cite{8759067}, which applies a classwise loss based on a normalized Gaussian distribution. DCWH introduces the class centers in the learning procedure with the essence of minimizing the Hamming distance between intra-class samples to the corresponding class center. 
Experiments show that DCWH outperforms prior pairwise/triplet labels-based deep hashing works especially on the datasets with a significant number of classes and small inter-class variations. Although DCWH guarantees that the average distance of intra-class samples to the corresponding class center is smaller than that to other centers. It still has its drawback. Concretely, when datapoints belonging to different classes are far from the corresponding class centers, they are prone to be closer to each other than their intra-class samples. We will demonstrate this issue in the next section in detail. To overcome these issues, we propose an improved deep classwise hashing method (IDCWH) with centers similarity learning to generate more compact and discriminative binary codes for image retrieval. The contributions of this paper are summarized as follows:
\begin{itemize}
  \item The proposed IDCWH enables networks to learn hashing codes and class centers simultaneously in an end-to-end manner. The dual part of the designed loss mutually contributes to the concentration of intra-class samples on the corresponding class center and it gives better training convergence.
  \item We introduce a two-step strategy on centers similarity learning. It first estimates binary centers of each unique class from mini-batch data to represent the intra-class samples. Then, it dynamically attracts the corresponding learnable center to be close to the estimated center while pushing other centers away in the Hamming space.
  \item Extensive experiments on three benchmark datasets show the proposed IDCWH outperforms other compared state-of-the-art baselines. It verifies the effectiveness of IDCWH on generating compact and semantics-preserving hashing codes for image retrieval.
\end{itemize}

\section{Related Works}
Denote the dataset with $N$ training samples as $X=\{x_i\}_{i=1}^N$. The label matrix of $X$ is given as $Y=[y_1,y_2,\ldots,y_N]\in\mathbb{R}^{C\times{N}}$, where $C$ is the number of unique classes in the dataset and ${y}_{ji}=1$ if $x_i$ owns the label of the $j$th class, otherwise $y_{ji}=0$. Note that, $y_i$ is not limited as a one-hot encoding vector, in order to handle the multi-label cases. The goal of supervised hashing is learning hashing functions to map $X$ into binary codes $B=\{b_i\}_{i=1}^N\in\mathbb{R}^{l\times{N}}$ with the supervision information in $Y$, where $b_i\in\{-1,1\}^l$ and $l$ is the code length. Nowadays, deep supervised hashing~\cite{li2017deep,li2016feature,8759067,chen2019deep} applies deep neural networks $f(\cdot)$ to achieve feature learning and hashing learning jointly. Thus, $b_i=f(\Theta;x_i)$, where $\Theta$ represents the learnable parameters of the network.

\subsection{Deep Supervised Hashing with Pairwise Labels}
Provided the label matrix $Y$, the pairwise similarity between any pair of samples $x_i$ and $x_j$ in $X$ can be represented as $s_{ij}$, where $s_{ij}=y_i^Ty_j$. Thus, $s_{ij}=1$ if $x_i$ and $x_j$ shares at least one common label, while $s_{ij}=0$ otherwise. Most deep supervised hashing is based on pairwise/triplet labels~\cite{li2016feature,li2017deep}, which learns $f(\cdot)$ to map $X$ from the original space into $B$ in Hamming space while maintaining the semantics similarity in $s_{ij}$. Thus, the loss function $L_p$ preserving the pairwise similarity is summarized as follow:
\begin{equation}
\min_\Theta L_p= \sum_{i=1 \atop i \neq j}^N\sum_{j=1}^NL(f(\Theta;x_i), f(\Theta;x_j);s_{ij})
\label{1}
\end{equation}

The above loss function expects to minimize the Hamming distance between any pair of similar samples involving $x_i$ while maximizing the distance between any pair of dissimilar samples involving $x_i$. Since the expensive computation cost of \eqref{1}, the pairwise-based deep supervised hashing usually samples a portion of data from the whole dataset for training. The insufficient utilization of the supervised information leads to the learned binary codes $B$ less discriminative for image retrieval.

\begin{figure}[b]
\centering
\includegraphics[width=0.4\textwidth]{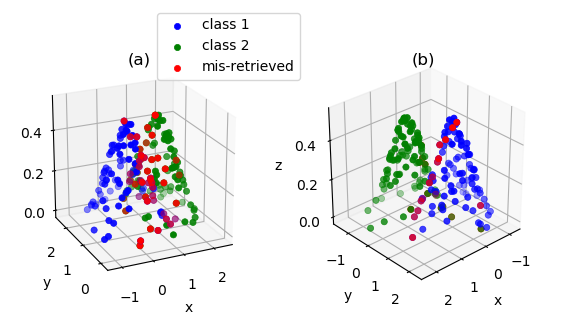}
\caption{A toy example demonstrates the drawback of original DCWH (a) and its improvement (b). Two classes of datapoints are sampled from two independent Gaussian distributions with the same variance under every case. The centers of two classes are located at (0,1) and (1,1) respectively in (a), whereas at (0,1) and (1,0) in (b). xy-axis represents the coordinates of data points while z-axis represents the value of probability density.}
\label{fig:1}
\end{figure}

\subsection{Deep Supervised Hashing with Classwise Loss}
Inspired from deep metric learning~\cite{oh2016deep}, DCWH~\cite{8759067} first applies a classwise loss for hashing learning, which directly uses the label information as supervision. The classwise loss can be formulated as a cross-entropy based normalized Gaussian distribution:
\begin{equation} \label{3}
\begin{split}
\min_{\Theta, M}L_1 &= -\sum_{i=1}^N\sum_{j=1}^Cy_{ji}\log{p\left(y_{ji}=1|x_i;\Theta;M\right)} \\
&= -\sum_{i=1}^{N} \sum_{j=1}^{C}y_{ji} \log \frac{\exp \left\{-\frac{\|h_{i}-\mu_{j}\|^2}{2 \sigma^{2}}\right\}}{\sum_{j=1}^{C} \exp \left\{-\frac{\|h_{i}-\mu_{j}\|^2}{2 \sigma^{2}}\right\}} 
\end{split}
\end{equation}
where $h_i=f(\Theta;x_i)$ is the output of feature learning part, $M =\{\mu_j\}_{j=1}^C$ are the class centers, and $\sigma^2$ is a hyperparameter controlling the class variance. Note the scale parameter $\frac{1}{\sqrt{2\pi}\sigma}$ is omitted for clarity, since it does not affect the final result. $M$ is updated periodically as $\mu_j=\left(\sum_{i=1}^{N}\mathrm{y}_{ji}h_i\right)/n_j$, where $n_j$ is the number of samples with the $j$th class label. By introducing class centers, \eqref{3} aims to maximize the probability of assigning intra-class samples to the corresponding class center. However, it cannot guarantee the samples belonging to different classes but lying far from the mean of each Gaussian distribution is closer to their intra-class samples. An illustrative example is shown in Fig.~\ref{fig:1}. We can see, increasing the Hamming distance between different class centers helps to avoid mis-retrieved datapoints lying on the boundary. We elaborate on how to improve this centers discriminability from next section.

\begin{figure*}[t]
\includegraphics[width=0.8\textwidth]{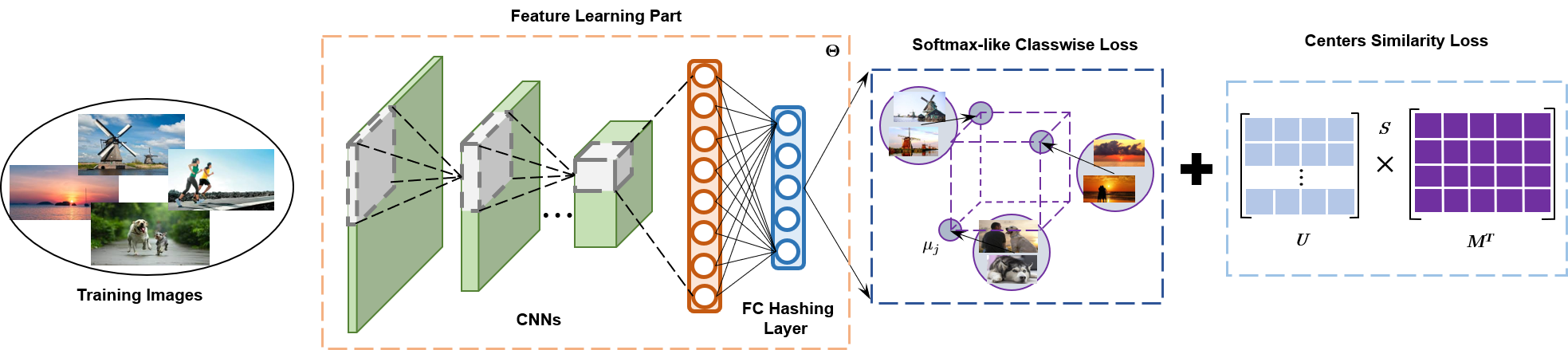}
\centering
\caption{Illustration of IDCWH. It learns the network parameters $\Theta$ in the feature learning part and the class centers $M=\{\mu_j\}$ in the dual loss part. The proposed centers similarity learning first estimates binary centers $U$, then centers similarity can be indicated by the inner product between $U$ and $M$ with the supervision of $S$.}
\label{fig:2}
\end{figure*}

\section{Improving Classwise Hashing with Centers Similarity Learning}
\subsection{Problem Definition}
To alleviate the aforementioned issue in DCWH, the proposed IDCWH introduces a novel loss function on centers similarity. For simplicity, we adopt the formulation of original classwise loss in \eqref{3} as the base loss and regard the newly designed loss as a regularization term. Note that, we now treat class centers as learnable parameters instead of being updated manually and periodically. IDCWH provides an end-to-end framework combining efficient hashing codes and class centers learning. Specially, 
we aim to learn the network parameters $\Theta$ in the hashing learning part and the class centers $M = \{\mu_j\}_{j=1}^C$ in the loss layer. The proposed centers similarity learning has two functionalities: firstly, it dynamically attracts the class centers to concentrate on intra-class samples, which interacts with the original classwise loss to minimize the samples-center distance. Secondly, it maximizes the distance between intra-class samples and centers belonging to other classes for discriminative retrieval. The overview of IDCWH is shown in Fig. \ref{fig:2}.

\subsection{Two-step Centers Similarity Learning}
\subsubsection{Intra-class Samples Clustering}
One straightforward way to improve the discriminability of the learnable centers is enlarging the Hamming distance between pairwise class centers. However, this may cause a problem that, the learned class centers cannot follow the distribution of the hashing outputs. 
Consequently, the training procedure may become unstable and hard to converge. To tackle this issue, we alternatively first estimate a center with binary constraint, which clusters the intra-class data points. Denote the approximated center of $j$th class as $u_j$, then we have the following objective function:
\begin{equation} \label{4}
\min_{u_j}\sum_{i=1}^N \sum_{j=1}^Cy_{ji}{\left\|u_j-b_i\right\|}_2^2 \quad s.t. \; u_j  \in\{-1, 1\}^l
\end{equation}
\eqref{4} can be reduced to the following formulation, which clusters intra-class samples to each approximated center individually:
\begin{equation} \label{5}
\begin{split}
 &\min_{u_j}\sum_{k=1}^{n_j}{\left\|u_j-b_{jk}\right\|}_2^2 \\
= \sum_{k=1}^{n_j} u_j^T &u_j + \sum_{k=1}^{n_j}b_{jk}^Tb_{jk} - 2u_j^T\sum_{k=1}^{n_j}b_{jk}
\end{split}
\end{equation}
where $n_j$ is the number of samples belonging to $j$th class and $b_{jk}$ represents the $k$th binary codes with the class label $j$. Since both $u_j^Tu_j$ and $b_{jk}^T b_{jk}$ are constants with the binary constraint, \eqref{5} can be simplified as:

\begin{equation} \label{6}
\max_{u_j}u_j^T \sum_{k=1}^{n_j}b_{jk} \quad s.t. \; u_j  \in\{-1, 1\}^l
\end{equation}

Denote the sum of all the intra-class binary codes $\sum_{k=1}^{n_j}b_{jk}$ as $m_j$, the solution of \eqref{6} is a closed-form equation shown as:
\begin{equation}
u_{j,v} = \mathbf{sgn}(m_j) = 
    \left\{
        \begin{array}{cc}
            1, &{m_{j,v} \geq 0} \\
           -1, &{m_{j,v}<0}
        \end{array}
    \right.
\label{7}
\end{equation}
where $\mathbf{sgn}(\cdot)$ is the binary function and $v=1,2,\ldots,l$ is the vector entry. \eqref{7} is similar to the process of voting for every bit of the binary center within the samples of the same class. 

\subsubsection{Centers Concentration and Repelling}After obtaining the $u_j$ of intra-class samples, the second step of the proposed method aims to minimize the Hamming distance from $u_j$ to the learnable class center $\mu_j$ while maximizing the distance from $u_j$ to other learnable class centers. Consider the training process is in mini-batch manner, we denote the estimated binary centers of all the unique labels appearing in mini-batch data as $U=\{u_i\}_{i=1}^Z$. Then, we can build a similarity matrix $S\in\mathbb{R}^{Z\times{C}}$ to connect the similarity relations between $U$ and $M$. $S$ is defined as follows:
\begin{equation}
s_{ij}  = 
    \left\{
        \begin{array}{cc}
            1, &{y_{u_i}^Ty_{\mu_j}=1} \\
            \\
            0, &{y_{u_i}^Ty_{\mu_j}=0}
        \end{array}
    \right.
\label{8}
\end{equation}
\eqref{8} means: if the learnable class center and the estimated binary center are within the same class, then their similarity is encoded as `1', otherwise `0'. Thus, our designed loss function should measure how well each pair of $u_i$ and $\mu_j$ matches given their similarity $s_{ij}$. 

Suppose both $u_i$ and $\mu_j$ are binary vectors, the Hamming distance between $u_i$ and $\mu_j$ is $0.5(l-u_i^T\mu_j)$. Let $\theta_{ij} = 0.5u_i^T\mu_j$ denotes half of the inner product of $u_i$ and $\mu_j$. Inspired from the widely-used likelihood in pairwise labels-based hashing~\cite{zhang2014supervised}, we present the likelihood of centers similarity $s_{ij}$ as following:
\begin{equation}
  p\left(s_{ij} | u_i;\mu_j \right)=
  \left\{
    \begin{array}{cc}
       \rho(\theta_{ij}),  & {s_{ij}=1} \\
        1-\rho(\theta_{ij}),  & {s_{ij}=0}
    \end{array}
    \right.
\label{9}
\end{equation}
where $\rho(\cdot)$ is sigmoid function. From \eqref{9}, it is clear that the larger the inner product of $u_i$ and $\mu_j$, the bigger the $p(1 |u_i,\mu_j)$ while the smaller the $p(0 |u_i,\mu_j)$. It corresponds with our expectation that the learnable class center is more likely to cluster intra-class samples while lying far away from the samples belonging to other classes. Noted that there is a binary assumption on $\mu_j$ in \eqref{9} while $\mu_j$ is a real-valued learnable parameter in implementation. Therefore, we replace $u_i^T\mu_j$ by its approximation with continuous relaxation. Observing that, there exists a nice relationship between $u_i^T\mu_j$ with binary constraint and their cosine similarity:
\begin{equation}
    u_i^T\mu_j= l \frac{u_i^T}{\|u_i\|} \frac{\mu_j}{\|\mu_j\|} = l \cos{\left(u_i, \mu_j\right)}
\label{10}
\end{equation}
where $\|\cdot\|$ is $\ell_2$ norm. Hence, we alternatively adopt $\theta_{ij} = 0.5 l \cos{(u_i, \mu_j)}$ in \eqref{9}. 

Based on \eqref{9}, we present the loss function of the proposed centers similarity learning, which is the negative log likelihood shown as:  
\begin{equation}
\begin{split}
 &\min_{M}L_2 = -\log \prod_{i=1,\ldots,Z \atop j=1,\ldots,C} p\left(s_{ij} | c_i;\mu_j \right) \\
 &=-\sum_{s_{ij} \in S}\left( s_{ij}\theta_{ij} - \log(1+e^{\theta_{ij}})\right)
\end{split}
\label{11}
\end{equation}
By adding \eqref{11} as a regularization term to \eqref{3}, we obtain the following loss function:
\begin{IEEEeqnarray*}{l}
\min_{\Theta, M} L_{1}+\gamma L_{2} = -\sum_{i=1}^{N} \sum_{j=1}^{C}y_{ji} \log \frac{\exp \left\{-\frac{\|h_{i}-\mu_{j}\|^2}{2 \sigma^{2}}\right\}}{\sum_{j=1}^{C} \exp \left\{-\frac{\|h_{i}- \mu_{j}\|^2}{2 \sigma^{2}}\right\}} \\ \qquad \qquad \qquad \quad \:
 - \gamma \sum_{s_{ij} \in S} \left( s_{ij}\theta_{ij} - \log(1+e^{\theta_{ij}}) \right) \IEEEyesnumber \\
 \IEEEeqnarraymulticol{1}{c}{
s.t.\quad h_{i}=b_{i}, \quad h_{i} \in \mathbb{R}^{l \times 1},\quad b_{i}=\{-1,1\}^{l} }
\end{IEEEeqnarray*}\label{12}
where $\gamma$ is a hyper-parameter balancing the class-wise loss $L_1$ and centers similarity loss $L_2$. 

Finally, we need to handle the discrete optimization of $h_i$. Follow the strategy in~\cite{li2016feature}, we relax $h_i$ from discrete to continuous, while introducing a quantization error between $h_i$ and $b_i=\mathbf{sgn}(h_i)$. By integrating with the quantization term, we obtain the finalized objective function shown as following:
\begin{equation}\label{13}
\begin{split}
 &\min_{\Theta,M}L = L_1 + \gamma L_2 + \beta \sum_{i=1}^{N}\left\|b_{i}-h_{i}\right\|_{2}^{2}
\end{split}
\end{equation}
where $\beta$ is a hyper-parameter controlling the weight of the quantization term.

\subsection{Implementation Details and Optimization}
Since we train the network in mini-batch iteration, to capture the information of all the intra-class samples in the hashing codes space, we dynamically estimate the binary centers $U=\{u_i\}_{i=1}^Z$ from the current binary sum and newly sampled data. Recall \eqref{7}, we denote the current binary sum of all the intra-class samples belonging to $j$th class as $\mathbf{sum}_j^{(t)}$. Then, for each iteration when new samples of $j$th class appear, $u_j$ is updated as follows:
\begin{equation} \label{14}
\begin{split}
    \mathbf{sum}_j^{(t+1)} &= \mathbf{sum}_j^{(t)} + m_j^{(t+1)} \\
    u_j^{(t+1)} &= \mathbf{sgn}(\mathbf{sum}_j^{(t+1)})
\end{split}
\end{equation}
where $m_j^{(t+1)}=\sum_{k=1}^{n_j}\mathbf{sgn}(h_{jk})$ represents the sum of binarized hashing outputs belonging to $j$th class in a newly-come mini-batch data. 

With each forward pass in the training procedure, we obtain the hashing outputs $\{h_i\}$ of the network and the estimated binary centers $U$. Then we take the derivatives of $L$ w.r.t. $h_i$ and $\mu_j$ individually and update $\mu_j$ by gradient descent, shown as:
\begin{equation}
\frac{\partial L}{\partial h_i} = \frac{\partial L_1}{\partial h_i}-2\beta(b_i-h_i); \quad  
\frac{\partial L}{\partial \mu_j} = \frac{\partial L_1}{\partial \mu_j}+\gamma \frac{\partial L_2}{\partial \mu_j}
\label{15}
\end{equation}
And ${\partial L_2}/{\partial \mu_j}$ can be derived as:
\begin{equation}
\frac{\partial L_2}{\partial \mu_j} = \frac{1}{2} \sum_{i=1}^Z u_i(r_{ij}-s_{ij})
\label{16}
\end{equation}
where $r_{ij}=\rho(0.5u_i^T\mu_j)$. Then, the gradients are backpropagated to the feature learning part to update network parameters $\Theta$. The overall learning procedure of IDCWH is summarized in Algorithm~\ref{alg:algorithm}.

\begin{algorithm}[tb]
\caption{Learning Algorithm for IDCWH}
\label{alg:algorithm}
\textbf{Input}: Training data $X=\{x_i\}_{i=1}^N$, label matrix $Y\in \mathbb{R}^{C\times{N}}$. \\
\textbf{Parameter}: Hyper parameters $\sigma^2$,$\gamma$ and $\beta$; code length $L$ and batch size $bs$.\\
\textbf{Output}: Hashing function $f(\cdot)$, class centers $M=\{\mu_j\}_{j=1}^C$.\\
Initialize network parameters $\Theta$ by Glorot distribution, class centers $M$ by normal distribution; epoch $T=0$.

\begin{algorithmic}[1] 
\STATE \textbf{Repeat}
\begin{ALC@g}
    \STATE $T=T+1$;
    \STATE $U = \mathbf{0}$
    \STATE \textbf{for} $\lfloor \frac{N}{bs} \rfloor$ iterations \textbf{do}
    \begin{ALC@g}
        \STATE Randomly sample a mini-batch data $\{x_i\}_{i=1}^{bs}$;
        \STATE Compute $h_i=f(\Theta;x_i)$ and $b_i = \mathbf{sgn}(h_i)$ by forward propagation;
        \STATE Update $\{u\}_{k=1}^Z$ for each unique class label appearing in the batch by \eqref{14};
        \STATE Compute the loss function $L$ in \eqref{13};
        \STATE Calculate derivatives of $L$ concerning $h_i$ and $\mu_j$ individually by \eqref{15} and \eqref{16};
        \STATE Update $M$ and $\Theta$ by back propagation;  \\
    \end{ALC@g}
    \STATE \textbf{end for}
\end{ALC@g}
\STATE \textbf{Until} Convergence
\end{algorithmic}
\end{algorithm}

\subsection{Out-of-Sample Extension}
After completing the training procedure, the learned network can generate binary hashing codes for unseen samples. We obtain the binary code $b_{test,i}$ of a given test sample $x_i$ by simply making a forward pass of the network and taking the binary operation, i.e.:
\begin{equation}
b_{test,i} = \mathbf{sgn}(f(\Theta;x_i))
\label{17}
\end{equation}

\section{Experiments}
To verify the superiority and effectiveness of our proposed IDCWH methods, we conduct extensive experiments on three large-scale benchmark datasets, i.e. CIFAR-10, CIFAR-100~\cite{krizhevsky2009learning} and MS-COCO~\cite{lin2014microsoft}. All the experiments are run by PyTorch with two Nvidia RTX2080 Ti GPU cards.

\subsection{Experiments Settings}
\textbf{CIFAR-10}~\cite{krizhevsky2009learning} is a standard dataset for image recognition with 60,000 images in 10 classes. The official split contains 50,000 training images with 5,000 images per class, and 10,000 testing images with 1,000 per class. We conduct two groups of experiments following different protocols as used in \cite{li2016feature,wang2016deep,li2017deep}. In the first protocol, we randomly choose 100 images per class (1000 images in total) as query set, and 500 images per class (5000 images in total) as training set. The rest is all used as the database. In the second protocol, all the images in the official training split are used for training and all the images in the test split are used for query. 

\textbf{CIFAR-100}~\cite{krizhevsky2009learning} is similar to CIFAR-10 except that it has 100 classes with 600 images per class. We follow the protocol as in DCWH~\cite{8759067}, which adopts the official training set and testing set of CIFAR-100. Specifically, we use 100 images per class (10,000 in total) as a query set and 500 images per class (50,000 in total) for training.

\textbf{MS-COCO}~\cite{lin2014microsoft} is a popular dataset for image recognition, object segmentation and captioning. The official release of `train2014' and `val2014' contains 82,783 training images and 40,504 validation images. Each image is labelled by some of the 80 semantic concepts. Follow the protocol in \cite{cao2017hashnet}, we combine the training and validation images together and randomly sample 5,000 images as query set and 10,000 images as a training set. The reminder images are all used as the database. 

We adopt the standard similarity protocol for evaluation as in prior works. Specifically, for the single-label datasets, i.e. CIFAR-10 and CIFAR-100, images from the same class are regarded as similar, otherwise dissimilar. For the multi-label dataset, i.e. MS-COCO, any pair of images sharing at least one common label are regarded as similar, otherwise dissimilar. In addition to Mean Average Precision (MAP), four other commonly-used evaluation metrics are used in experiments, i.e. Precision-Recall curves (PR), Precision curve within two Hamming distances w.r.t. different code lengths (P@H=2), Precision curves w.r.t. different number of top returned images (P@N) and Recall curves with two Hamming distances w.r.t. different code lengths (R@H=2). 

We compare our proposed method with original DCWH~\cite{8759067} and several classical deep supervised hashing methods including DPSH~\cite{li2016feature}, DTSH~\cite{wang2016deep}, DSDH~\cite{li2017deep} and HashNet~\cite{cao2017hashnet}. Besides, we also compare the performance with traditional supervised methods i.e. SDH~\cite{shen2015supervised} and FSDH~\cite{gui2017fast}, combined with deep feature learning. The two methods are denoted as SDH+CNN and FSDH+CNN. For the feature learning part, we simply employ GoogLeNet~\cite{szegedy2015going} with batch normalization~\cite{ioffe2015batch} as the backbone network. The network has been pretrained on ImageNet. Since some other works apply the different network architecture with us, for the fair comparison, we re-run the experiments with the codes released by authors and we report performances of both the original works and their counterparts with GoogLeNet. The latter are marked with an asterisk, e.g. DPSH* and DSDH*. 

The configurations of proposed IDCWH are shown as follows: we finetune the pretrianed GoogLeNet from the input to the last layer before the fully-connected layer and train the FC hashing layer (as shown in Fig.~\ref{fig:2}) for 150 epochs. We use mini-batch stochastic gradient descent (SGD) with momentum 0.9 and weight decay 5e-4. The initial learning rates are 1e-2 and 5e-3 for feature learning part and centers learning, respectively. And the learning rates are decayed by 0.1 every 50 epochs. The three hyperparameters adopted by cross-validation are $\sigma^2=4$, $\gamma=1$ and $\beta=0.01$. The batch size is fixed to 128 during all the experiments.

\begin{table*}[t]
    \centering
    \small
    \caption{MAP results by Hamming ranking on CIFAR-10 under two experiment protocols}
    \begin{tabular}{ccccccccc}
        \toprule
        \multirow{2}{*}{Method} & \multicolumn{4}{c}{CIFAR-10 (mini)} & \multicolumn{4}{c}{CIFAR-10 (full)} \\
        \cmidrule(lr){2-5}\cmidrule(lr){6-9}
        &\textbf{12 bits} &\textbf{24 bits} &\textbf{32 bits} &\textbf{48 bits} &\textbf{12 bits} &\textbf{24 bits} &\textbf{32 bits} &\textbf{48 bits} \\
        \midrule
        SDH+CNN  &0.207  &0.218  &0.223  &0.210  &0.364  &0.433  &0.405  &0.414 \\
        FSDH+CNN  &0.196  &0.220  &0.203  &0.212 &0.374  &0.443  &0.410  &0.446 \\
        DPSH &0.713 &0.727 &0.744 &0.757 &0.763 &0.781 &0.795 &0.807 \\
        DPSH*  &0.797  &0.806  &0.820  &0.802  &0.908  &0.922  &0.925  &0.935 \\
        DTSH   &0.710  &0.750  &0.765  &0.774  &0.915 &0.923  &0.925  &0.926 \\
        DTSH* &0.790 &0.797 &0.794 &0.775 &0.928 &0.935 &0.940 &0.942 \\
        DSDH  &0.740  &0.786  &0.801  &0.820  &0.935  &0.940  &0.939  &0.939 \\
        DSDH* &0.800 &0.802 &0.804 &0.808 &0.913 &0.925 &0.943 &0.930 \\
        DCWH  &0.818  &0.840  &0.848  &0.850  &0.940  &0.950  &0.954  &0.952 \\
        IDCWH  &$\mathbf{0.828}$  &$\mathbf{0.865}$  &$\mathbf{0.868}$  &$\mathbf{0.859}$ &$\mathbf{0.964}$  &$\mathbf{0.969}$  &$\mathbf{0.967}$  &$\mathbf{0.968}$ \\
        \bottomrule
    \end{tabular}
    \label{table:map1}
\end{table*}

\begin{figure*}[t]
\includegraphics[width=0.85\textwidth]{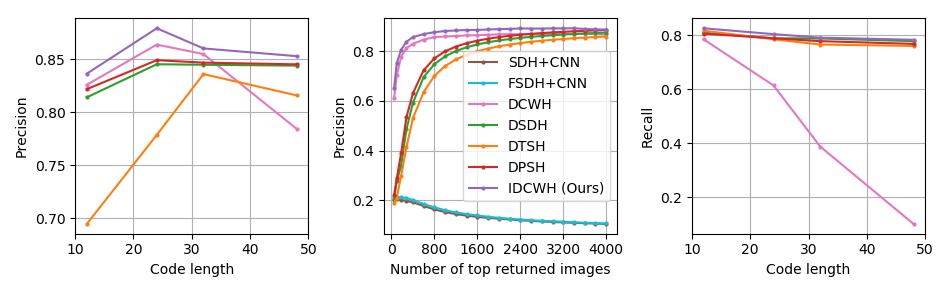}
\centering
\caption{Results on CIFAR-10 dataset. Left: P@H=2 w.r.t. different code lengths; Middle: P@N under 32-bit codes; Right: R@H=2 w.r.t. different code lengths.}
\label{fig:3}
\end{figure*}

\subsection{Results}
We first present the evaluation on CIFAR-10 dataset. Denote the first protocol and second protocol of the experiments as CIFAR-10 (mini) and CIFAR-10 (full), respectively. The MAP results of compared methods are shown in Table \ref{table:map1}. We can see that the proposed IDCWH outperforms other methods under almost all the cases of two experiment protocols. Among them, the classwise-based deep hashing methods perform better than other pairwise/triplet labels-based methods. And IDCWH surpasses the performance of DCWH by 1-2\%. Specifically, in the first protocol, it achieves the best MAP performance with 86.8\% under 32-bit codes. And the best MAP result on the second protocol is 96.9\% under 24-bit codes. One can find, DPSH* and DTSH* with employed GoogLeNet backbone, perform significantly better than the original works with their counterpart networks. While the performance of the combination methods of traditional supervised hashing and deep feature learning, i.e., SDH+CNN and FSDH+CNN, is really poor. It verifies the superiority of end-to-end training manner of deep supervised hashing. One can also observe, by training with more samples, the performances on the second protocol of all the methods are greatly improved compared to those of the first protocol. 

The performance on CIFAR-10 under other evaluation metrics is shown in Fig. \ref{fig:3}. It is clear that the proposed IDCWH is robust as it achieves the best result under various evaluation metrics. From Fig. \ref{fig:3}, one can observe two common phenomenons from all the compared methods. The first is, the curve of P@H=2 always first increases then drops with the growth of code length. The second is the overall trend of R@H=2 is decreasing with the increase in code length. The reason may be explained as: with the dimension of Hamming space, the learned binary codes become sparser, which leads to fewer data points fallen into the range within two Hamming distances. 

\begin{table}[t]
    \centering
    \caption{MAP results by Hamming ranking on CIFAR-100 dataset}
    \begin{tabular}{ccccc}
        \toprule
        \multirow{2}{*}{Method} & \multicolumn{4}{c}{CIFAR-100} \\
        \cmidrule(lr){2-5}
        &\textbf{12 bits} &\textbf{24 bits} &\textbf{32 bits} &\textbf{48 bits}\\
        \midrule
        SDH+CNN  &0.0617  &0.0624  &0.0610  &0.0668   \\
        FSDH+CNN  &0.0596  &0.0618  &0.0650  &0.0665  \\
        DPSH &0.0597 &0.1008 &0.1196 &0.1587  \\
        DTSH   &0.6070  &0.7056  &0.7122  &0.7252   \\
        DSDH  &0.0784  &0.1495  &0.1868  &0.2272  \\
        DCWH  &0.7227  &0.7441  &0.7570  &0.7658   \\
        IDCWH  &$\mathbf{0.7642}$  &$\mathbf{0.8130}$  &$\mathbf{0.8236}$  &$\mathbf{0.8351}$\\
        \bottomrule
    \end{tabular}
    \label{table:cifar100}
\end{table}

We further present the experimental results on CIFAR-100 dataset. The performance of the MAP on CIFAR-100 is presented in Table \ref{table:cifar100}. We can see the proposed IDCWH achieves the best results under all the compared code lengths and it surpasses the second place, i.e. DCWH with an at least 4\% margin. Specifically, it greatly improves the MAP scores to more than 80\% with codes length not less than 24-bit, which is nearly 7\% higher than DCWH under all the cases. While for 12-bit codes, it outperforms other methods with 76.42\% MAP performance. 

\begin{figure*}[t]
\includegraphics[width=0.85\textwidth]{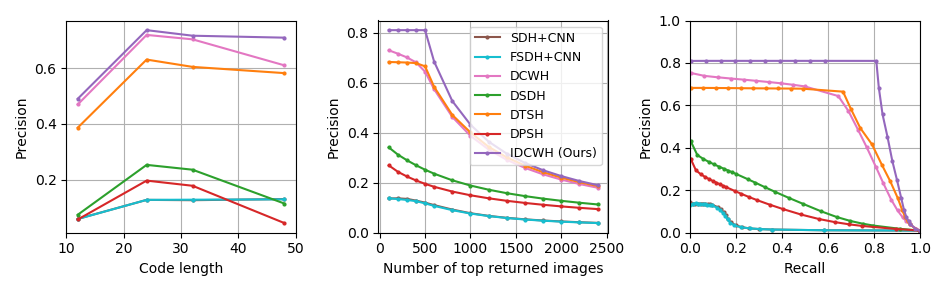}
\centering
\caption{Results on CIFAR-100 dataset. Left: P@H=2 w.r.t. different code lengths; Middle: P@N with 48-bit codes; Right: PR curve with 48-bit codes.}
\label{fig:4}
\end{figure*}

For other pairwise/triplet labels-based methods, the performance on CIFAR-100 is much poorer than that on CIFAR-10.  However, IDCWH and DCWH, both of which are based on classwise loss, can still achieve promising results. Consider the increase in the number of classes and the less number of training images in CIFAR-100. It verifies the effectiveness of classwise hashing, which can better capture the semantics information for a large-scale dataset with small inter-class variations and significant intra-class variations. The performance on other evaluation metrics, i.e. P@2, P@N and PR curves, are illustrated in Fig. \ref{fig:4}. We can see the proposed IDCWH again preforms the best under various evaluation metrics. Observe that, there is a drastic drop on the curve of P@N when N=500, this corresponds with the fact that the number of training images per class in CIFAR-100 is 500. One can also find that, among all the pairwise/triplet labels-based methods, DTSH performs much better than DPSH and DSDH.

Finally, we summarize the evaluation on the multi-labeled MS-COCO dataset. The performance with HashNet, DHN~\cite{zhu2016deep} and DNNH~\cite{lai2015simultaneous} are added, which are directly taken in \cite{cao2017hashnet} and we also run experiments with DSDH and DPSH methods. The MAP result of all the compared methods on MS-COCO is presented in Table~\ref{table:coco}. We can see IDCWH performs slightly better than DCWH under each code length and it achieves the highest MAP scores among all the methods. Recall the formulation of IDCWH, it treats each class center independently as learnable hyperparameters instead of updating the center manually. Thus, it can be extended to multi-label cases feasibly without computing the average of any new combination of multi-class centers like in DCWH. One can also observe the performances of DPSH and DSDH are poor on MS-COCO. It again validates our expectation that, by leveraging the classwise difference between all the samples, it helps to generate more semantics-preserving binary codes. 

\begin{table}[t]
    \centering
    \caption{MAP results by Hamming ranking on MS-COCO dataset}
    \begin{tabular}{ccccc}
        \toprule
        \multirow{2}{*}{Method} & \multicolumn{4}{c}{MS-COCO} \\
        \cmidrule(lr){2-5}
        &\textbf{16 bits} &\textbf{32 bits} &\textbf{48 bits} &\textbf{64 bits}\\
        \midrule
        DPSH &0.3493 &0.3545 &0.3595 &0.3670  \\
        DSDH  &0.3470  &0.3587  &0.3661  &0.3703  \\
        DNNH &0.5932 &0.6034 &0.6045 &0.6099 \\
        DHN &0.6774 &0.7013 &0.6948 &0.6944 \\
        HashNet &0.6873 &0.7184 &0.7301 &0.7362 \\
        DCWH  &0.7227  &0.7441  &0.7570  &0.7658   \\
        IDCWH  &$\mathbf{0.7321}$  &$\mathbf{0.7597}$  &$\mathbf{0.7636}$  &$\mathbf{0.7698}$\\
        \bottomrule
    \end{tabular}
    \label{table:coco}
\end{table}

\subsection{Empirical Analysis}
\subsubsection{Visualization of Hashing Codes}
\begin{figure}[t]
\centering
\includegraphics[width=0.45\textwidth]{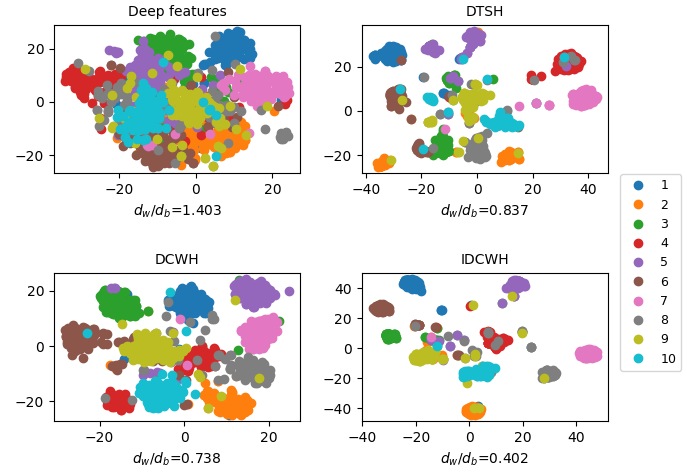}
\caption{The t-SNE visualization of hashing codes learned from deep features, DTSH, DCWH and the proposed IDCWH, respectively. $d_w$ and $d_b$ represent the within-class distance and the between-class distances.}
\label{fig:5}
\end{figure}

To intuitively compare the capability of different methods, we visualize the generated 48-bit hashing codes by DTSH, DCWH and IDCWH on CIFAR-100 dataset. For a more comprehensive comparison, we also visualize the original 1024-dimensional features extracted from deep CNNs. We first randomly sample 10 categories out of 100 categories in CIFAR-100, then generate the binary codes of all the testing images belonging to the 10 categories. Finally, t-SNE~\cite{maaten2008visualizing} is used to map the 48-bit codes/ 1024-dimensional features to 2-dimensional features for visualization. The results are shown in Fig. \ref{fig:5}. We observe that the hashing codes generated by IDCWH are most compact and discriminative. While the hashing codes generated by DTSH and DCWH are not well separated, which also reflects in the ratio of within-class distances and between-class distances of each method. 

\subsubsection{Ablation Study}
We conduct experiments to investigate the effect of the proposed centers similarity learning. Denote the finalized loss function without using $L_2$ shown in \eqref{11} as IDCWH-Single. The performance comparison of two configurations on three datasets is shown in Table \ref{table:ablation}. We can see IDCWH steadily performs better than IDCWH-Single under all the cases. It validates the functionality of centers similarity learning, which contributes to make the class center more concentrate on the intra-class samples while pushing other class centers as far as possible. Another observation from Table ~\ref{table:ablation} is, the contribution of $L_2$ is more distinct on shorter code lengths than the relatively longer code lengths.
\begin{table*}[t]
    \centering
    \small
    \caption{Comparison on MAP with single classwise loss and the combination with centers similarity loss}
    \setlength{\tabcolsep}{1.1mm}
    \begin{tabular}{ccccccccccccc}
        \toprule
        \multirow{2}{*}{Method} & \multicolumn{4}{c}{CIFAR-10 (mini)} & \multicolumn{4}{c}{CIFAR-100} & \multicolumn{4}{c}{MS-COCO} \\
        \cmidrule(lr){2-5}\cmidrule(lr){6-9} \cmidrule(lr){10-13}
        & 12-bit &24-bit &32-bit &48-bit &12-bit &24-bit &32-bit &48-bit &16-bit &32-bit &48-bit &64-bit \\
        \midrule
        IDCWH-Single  &0.7944  &0.8239  &0.8416  &0.8420  &0.7015  &0.7393  &0.7638  &0.7701 &0.6963 &0.7125 &0.7490 &0.7529\\
        IDCWH  &$\mathbf{0.8284}$  &$\mathbf{0.8650}$  &$\mathbf{0.8681}$  &$\mathbf{0.8586}$ &$\mathbf{0.7642}$  &$\mathbf{0.8130}$  &$\mathbf{0.8236}$  &$\mathbf{0.8351}$ &$\mathbf{0.7321}$  &$\mathbf{0.7597}$  &$\mathbf{0.7636}$  &$\mathbf{0.7698}$\\
        \bottomrule
    \end{tabular}
    \label{table:ablation}
\end{table*}

\subsubsection{Parameters Sensitivity Analysis}
\begin{figure}[t]
\centering
\includegraphics[width=0.45\textwidth]{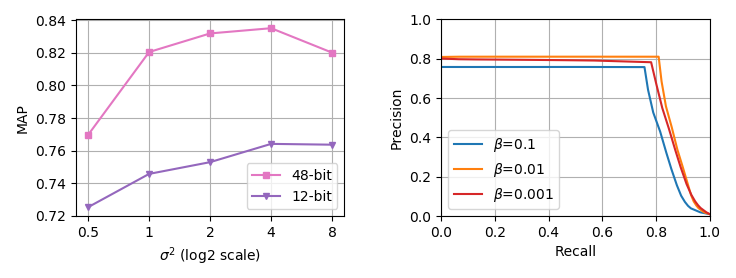}
\caption{Parameters sensitivity analysis. Left: MAP w.r.t. various $\sigma^2$ in 48-bit and 12-bit codes, respectively. Right: PR curve w.r.t. different values of $\beta$ in 48-bit codes.}
\label{fig:6}
\end{figure}

We further present the investigation on the influence of hyperparameters on the IDCWH. We focus on two hyperparameters in our work, i.e. $\sigma^2$ which controls the variance of intra-class samples to the corresponding class center and $\beta$ which penalizes the quantization error. Specifically, we test various $\sigma^2$ ranging within $\{0.5, 1, 2, 4, 8\}$ and $\beta$ ranging with $\{0.001, 0.01, 0.1\}$. The experiments are conducted on CIFAR-100 dataset and the result is illustrated in Fig. \ref{fig:6}. From Fig. \ref{fig:6}, we observe the performances of $\sigma^2$ and $\beta$ share similar patterns with the growth on the values, i.e. first rises then descends. Thus, we choose $\sigma^2=4$ and $\beta=0.01$ in this paper, which should be a promising configuration.

\section{Conclusion}
This paper addressed the limitation of prior works on deep supervised hashing and proposed an improved framework with centers similarity learning. To achieve this, firstly, it clusters intra-class samples to the learnable class centers and dynamically estimates binary centers within each iteration. To enhance the hashing codes discriminativity, it then minimizes the Hamming distance from the learnable class center to the corresponding estimated binary center while maximizing other estimated centers' distance. Experiments on three benchmark datasets verify that the proposed method can effectively boost original deep classwise hashing and yield state-of-the-art retrieval results. 
\section*{Acknowledgment}
This work is supported by Hong Kong Research Grants Council (Project C1007-15G) and City University of Hong Kong (Projects 9610034 and 9610460).

\bibliographystyle{IEEEtran}
\bibliography{conference.bib}

\end{document}